%% file: main.tex
\documentclass[10pt,twocolumn,letterpaper]{article}

\usepackage{cvpr}              
\input{preamble}
\definecolor{cvprblue}{rgb}{0.21,0.49,0.74}
\usepackage[pagebackref,breaklinks,colorlinks,allcolors=cvprblue]{hyperref}
\usepackage[accsupp]{axessibility}  

\usepackage{multirow}
\usepackage{colortbl}
\usepackage{hhline}
\usepackage{enumitem}
\usepackage{pifont}

\definecolor{red}{RGB}{255,0,0}
\definecolor{green}{RGB}{0,255,0}
\definecolor{blue}{RGB}{65, 105, 225}
\definecolor{orange}{RGB}{255,165,0}
\definecolor{title_gray}{gray}{.9}

\newcommand{\thickhline}{\noalign{\hrule height 1pt}}

\def\paperID{34633} 
\def\confName{CVPR}
\def\confYear{2026}

\title{Geometry-Guided 3D Visual Token Pruning for Video-Language Models}

\author{
Han Li \\
Beihang University \\
Zhongguancun Academy \\
{\tt\small lihan0620@buaa.edu.cn}
\and
Zehao Huang \\
{\tt\small zehaohuang18@gmail.com}
\and
Jiahui Fu \\
Beihang University \\
{\tt\small jiahuifu@buaa.edu.cn}
\and
Naiyan Wang \\
{\tt\small winsty@gmail.com}
\and
Si Liu \\
Beihang University \\
{\tt\small liusi@buaa.edu.cn}
}

\begin{document}
\maketitle


\input{sec/0_abstract}    
\input{sec/1_introduction}
\input{sec/2_related_work}
\input{sec/3_method}
\input{sec/4_experiments}
\input{sec/5_conclusion}

\section*{Acknowledgements}
This research is supported in part by the Zhongguancun Academy (Grant No. 20240304), National Key R\&D Program of China (2022ZD0115502), National Natural Science Foundation of China (No. 62461160308, No. 62576024, U23B2010), ``the Fundamental Research Funds for the Central Universities'' (No. 501RCQD2025), ``Pioneer'' and ``Leading Goose'' R\&D Program of Zhejiang (No. 2024C01161), Beijing Natural Science Foundation (QY25227), Ningbo Science and Technology Innovation 2025 Major Project (2025Z034), NSFCRGC Project (N CUHK498/24).

{
    \small
    \bibliographystyle{ieeenat_fullname}
    \bibliography{main}
}

\input{sec/X_suppl}

\end{document}

%% file: sec/0_abstract.tex
\begin{abstract}
Multimodal large language models have demonstrated remarkable capabilities in 2D vision, motivating their extension to 3D scene understanding. Recent studies represent 3D scenes as 3D spatial videos composed of image sequences with depth and camera pose information, enabling pre-trained video-language models to perform 3D reasoning tasks.
However, the large number of visual tokens in spatial videos remains a major bottleneck for efficient inference and context management. Existing pruning methods overlook the view consistency of spatial videos and the spatial diversity of the remaining tokens, which prevents them from effectively removing inter-frame redundancy and preserving scene completeness.
In this paper, we propose \textbf{Geo3DPruner}, a \textbf{Geo}metry-Guided \textbf{3D} Visual Token \textbf{Prun}ing framework. Geo3DPruner first models cross-frame relevance through geometry-aware global attention, and then performs a two-stage pruning process. The intra-voxel stage selects representative multi-view features within each voxel, while the inter-voxel stage preserves spatial diversity by selecting a globally distributed subset of voxels.
Extensive experiments on multiple 3D scene understanding benchmarks demonstrate that Geo3DPruner retains over 90\% of the original performance while pruning 90\% of visual tokens, significantly outperforming existing text-guided and vision-guided pruning methods.
The code will be released at \href{https://github.com/homothetic/Geo3DPruner}{https://github.com/homothetic/Geo3DPruner}.
\end{abstract}

%% file: sec/1_introduction.tex
\section{Introduction}
Empowering multimodal large language models~(MLLMs) with the ability to understand 3D scenes represents a critical milestone toward realizing spatial intelligence in the physical world.
However, the scarcity of large-scale 3D data representations~(\eg, point clouds, meshes) remains a fundamental bottleneck that hinders further progress.
To mitigate this limitation, recent studies~\cite{Video3DLLM, GPT4Scene, VGLLM} reformulate 3D scenes as 3D spatial videos, defined as sequences of image frames augmented with spatial cues such as camera poses and depth information.
This formulation allows video-language models~(VideoLMs) to leverage the 2D visual knowledge acquired through pre-training on large-scale image-text corpora, thereby supporting effective learning of 3D spatial structures.

\begin{figure}[t]
  \centering
  \includegraphics[width=1.0\linewidth]{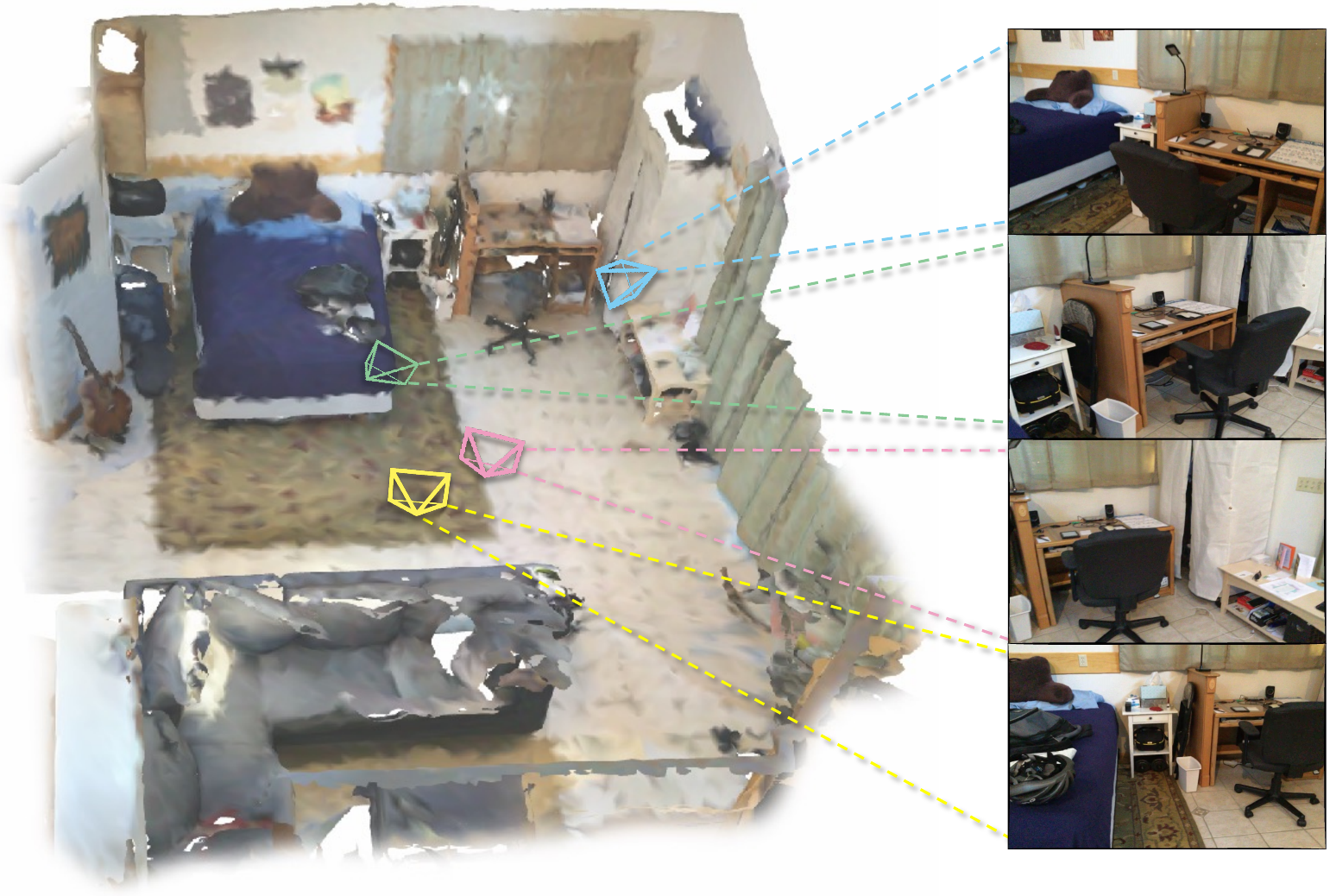}
  \caption{\textbf{Motivation of Geo3DPruner.} 
  3D spatial videos essentially represent multi-view projections of the complete 3D scene. Features corresponding to the same objects~(\eg, the wooden desk or the swivel chair) frequently recur across different frames. Existing pruning strategies fail to remove such redundancy due to the absence of global cross-frame relevance modeling.
  }
  \vspace{-5mm}
  \label{fig:motivation}
\end{figure}

Nonetheless, video-based representations remain constrained by the large number of visual tokens.
Increasing the number of sampled frames or employing higher-resolution images to capture more comprehensive 3D scene information substantially elevates computational demands, thereby significantly reducing inference efficiency.
To address this challenge, prior studies have proposed a variety of visual token pruning strategies, which can be broadly categorized into:
(1)~Text-Guided Pruning~\cite{FastV, SparseVLM, PyramidDrop, FitPrune}, which exploits cross-attention between textual and visual features to select question-relevant visual tokens;
(2)~Vision-Guided Pruning~\cite{VisPruner, VTC-CLS, VisionZip, SGL, LOOK-M}, which relies exclusively on visual features, typically identifying salient tokens via cross-attention with a [\texttt{CLS}] token or through internal self-attention mechanisms.

Although these methods have made notable progress in reducing visual token counts for general-purpose videos, their operations are typically confined to individual frames or temporally adjacent frames.
Given that 3D spatial videos essentially represent multi-view projections of the complete 3D scene, such temporally constrained pruning strategies fail to capture the intrinsic \textit{view consistency}, where features corresponding to the same spatial location may recur across arbitrary frames.
As illustrated in~\cref{fig:motivation}, objects such as the wooden desk and the swivel chair appear in multiple viewpoints, leading to substantial redundancy of features across frames. Existing pruning methods struggle to effectively remove this global redundancy.
Moreover, conventional pruning metrics predominantly focus on visual salience or text relevance, overlooking the \textit{spatial diversity} of retained tokens, \ie, ensuring that tokens are distributed across different objects and spatial regions within the scene. This property is essential for object-centric 3D tasks, including 3D dense captioning and 3D visual grounding.

To address these challenges, we propose \textbf{Geo3DPruner}, a \textbf{Geo}metry-Guided \textbf{3D} Visual Token \textbf{Prun}ing framework.
Unlike existing methods, Geo3DPruner first leverages geometric relevance across inter-frame features to construct a global cross-frame attention map.
Simultaneously, it aligns 2D visual features with their corresponding 3D voxel representations and employs a two-stage pruning strategy: intra-voxel view consistency pruning, which evaluates the contribution of multi-view features within each voxel to identify key tokens; and inter-voxel spatial diversity pruning, which measures correlations across voxels to select tokens that ensure comprehensive coverage of the scene.
This two-stage pruning process effectively removes redundant features arising from varied viewpoints while preserving the overall spatial completeness of the 3D scene.

Specifically, input video frames are processed through two parallel encoder branches: a 2D vision encoder~(\eg, SigLIP~\cite{SigLIP}) extracts 2D image features, while a 3D geometry encoder~(\eg, VGGT~\cite{VGGT}) captures 3D geometric features.
Within the geometry encoding branch, a global attention mechanism models long-range dependencies across inter-frame features, effectively capturing cross-view geometric consistency and enabling joint prediction of camera parameters and depth maps.
Conditioned on the reconstructed 3D geometry, 2D visual features are assigned to their corresponding voxel locations, establishing the foundation for the subsequent two-stage pruning process.
In the first stage, intra-voxel view consistency pruning evaluates the contribution of multi-view features within each voxel based on attention scores among features from the same spatial location across different viewpoints, identifying the most representative visual tokens and effectively aggregating observations from multiple views.
In the second stage, inter-voxel spatial diversity pruning measures correlations among voxel features via attention scores across different spatial locations and selects a geometrically diverse subset of voxels, ensuring comprehensive coverage while maintaining a compact 3D scene representation.
We conducted a comprehensive evaluation of Geo3DPruner on several 3D scene understanding benchmarks, including ScanRefer~\cite{ScanRefer}, Multi3DRefer~\cite{Multi3DRefer}, Scan2Cap~\cite{Scan2Cap}, ScanQA~\cite{ScanQA}, and SQA3D~\cite{SQA3D}.
The experimental results demonstrate that Geo3DPruner substantially reduces the number of visual tokens while preserving task performance, thereby significantly enhancing the inference efficiency of VideoLMs.

Our contributions are summarized as threefold:
\begin{itemize}
\item We introduce Geo3DPruner, a geometry-guided 3D visual token pruning framework that leverages global cross-frame relevance to eliminate visual redundancy arising from different viewpoints, while preserving the spatial completeness of 3D scenes.
\item We propose a two-stage pruning strategy. In the intra-voxel stage, the most representative tokens are selected by evaluating multi-view contributions within each voxel; in the inter-voxel stage, correlations across voxels are used to retain a geometrically diverse set of tokens.
\item We conduct extensive experiments on multiple 3D scene understanding benchmarks. The results demonstrate that Geo3DPruner substantially improves inference efficiency while maintaining strong task performance.
\end{itemize}

%% file: sec/2_related_work.tex
\section{Related Work}

\begin{figure*}[t]
  \centering
  \includegraphics[width=0.9\linewidth]{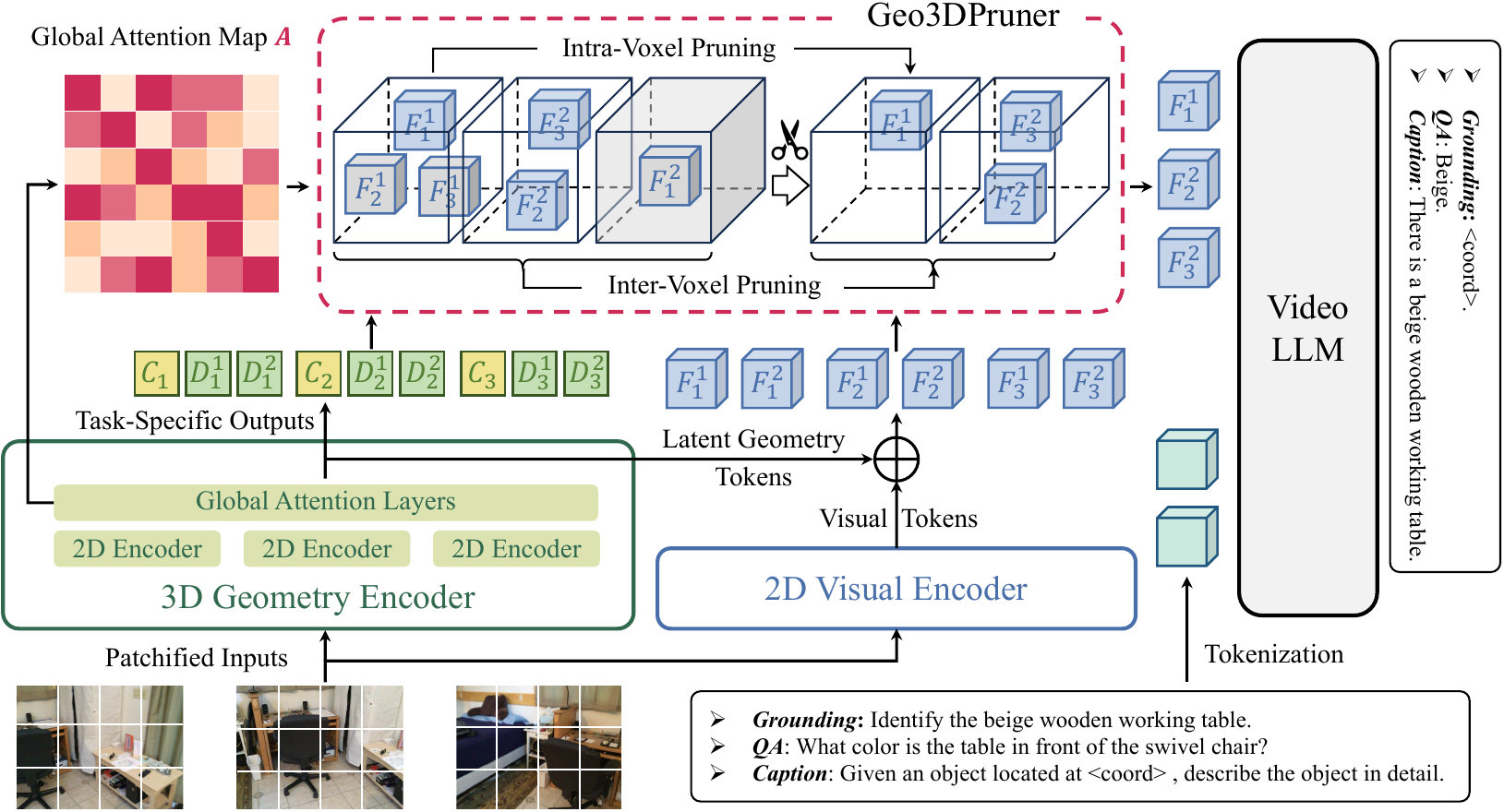}
  \caption{
  \textbf{Framework of Geo3DPruner.}
  We adopt Video-3D LLM~\cite{Video3DLLM} as our base model and replace its 3D positional encodings with geometry features following VG LLM~\cite{VGLLM}.
  Input video frames are processed by two parallel encoders: a 2D visual encoder~(\eg, SigLIP~\cite{SigLIP}) extracts image features, and a 3D geometry encoder~(\eg, VGGT~\cite{VGGT}) captures geometric features while modeling long-range cross-frame dependencies. The geometry-augmented visual features are assigned to voxel locations based on the reconstructed 3D geometry. The two-stage pruning process first selects representative multi-view tokens within each voxel and then chooses a spatially diverse subset of voxels, effectively removing redundancy while preserving the structural integrity of the 3D scene.
  }
  \vspace{-5mm}
  \label{fig:framework}
\end{figure*}

\subsection{MLLMs for 3D Scene Understanding}
Recent efforts have increasingly explored leveraging MLLMs for 3D scene understanding~\cite{ScanRefer, 3DSPS, Scan2Cap, ScanQA, EmbodiedScan, ECFusion, VSIBench}. Early methods in this direction~\cite{3D-LLM, PointLLM, PointLLM-V2, LL3DA, LEO, LLaVA-3D} typically transform 3D scenes into point cloud-based or voxel-based representations.
PointLLM~\cite{PointLLM, PointLLM-V2} couples a point cloud encoder with a powerful LLM to enable fine-grained, point-level understanding.
LLaVA-3D~\cite{LLaVA-3D} introduces a 3D patch representation that aggregates 2D patch features within voxel space to facilitate multi-view fusion.
Despite this progress, these methods suffer from a fundamental mismatch with pretrained MLLMs, which are primarily trained on large-scale 2D image datasets. 
To mitigate this gap, a growing line of work~\cite{Video3DLLM, GPT4Scene, VGLLM} explores using VideoLMs for 3D scene understanding.
Video-3D LLM~\cite{Video3DLLM} integrates 3D positional embeddings directly into video features to encode spatial cues.
VG LLM~\cite{VGLLM} further introduces a 3D geometry encoder to extract spatial priors from videos and injects them into the visual tokens.

\subsection{Visual Token Pruning}
One promising direction for accelerating MLLMs inference is to reduce the number of visual tokens. Earlier methods~\cite{LLaMA-VID, LLaVA-Mini, LLaVA-PruMerge} often require additional fine-tuning of the base model, whereas more recent training-free techniques can be directly applied to pretrained MLLMs without any parameter updates.
FastV~\cite{FastV} proposes a simple strategy that removes tokens with low attention scores to the final text tokens after the second Transformer~\cite{Transformer} layer. Building on this idea, a growing body of work~\cite{SparseVLM, PyramidDrop, FitPrune} leverages text-vision attention within the language model to estimate token importance.
Another line of research~\cite{VisPruner, VTC-CLS, VisionZip, SGL, LOOK-M} exploits visual cues for more effective token reduction. Among these, VisPruner~\cite{VisPruner} selects a compact subset of important tokens based on visual attention and further eliminates redundant ones by measuring inter-token similarity.
Compared with images, videos contain substantially more redundancy, making visual token reduction considerably more challenging for VideoLMs~\cite{PruneVid, KVTP, DyCoke}.
For example, PruneVid~\cite{PruneVid} merges spatiotemporal tokens based on their relevance to the question tokens.
Despite these advances, limited research has explored visual token pruning for VideoLMs in 3D scene understanding tasks.

\subsection{Feed-Forward 3D Reconstruction}
Traditional 3D reconstruction pipelines~\cite{SFM, 3DGS} typically rely on per-scene optimization, leading to high computational cost and limited scalability in real-world applications. 
To overcome this limitation, feed-forward reconstruction paradigms have emerged, aiming to infer scene geometry through a single network forward pass. 
Early efforts in this direction include learning-based multi-view stereo methods~\cite{MVSNet, CasMVSNet} and generalizable neural rendering methods~\cite{pixelNeRF, IBRNet}.
Building upon these developments, subsequent research~\cite{DUSt3R, MASt3R, Fast3R, MUSt3R, CUT3R, VGGT, StreamVGGT} introduces pointmap-based representations that explicitly encode pixel-aligned 3D geometry and cross-view correspondences within a unified framework. 
DUSt3R~\cite{DUSt3R} trains a Transformer-based~\cite{Transformer} encoder-decoder to directly predict two pixel-aligned pointmaps from an image pair.
To handle multi-view cameras, Fast3R~\cite{Fast3R} introduces a global-fusion Transformer~\cite{Transformer} that processes multiple views simultaneously.
VGGT~\cite{VGGT} presents a large Transformer-based architecture that directly predicts essential 3D attributes, achieving state-of-the-art performance in point reconstruction.
In this work, we leverage geometry features encoded by 3D reconstruction models to model global cross-frame relevance and perform 3D visual token pruning.

%% file: sec/3_method.tex
\section{Method}
To address the limitations of existing visual token pruning methods in VideoLMs for 3D scene understanding, we introduce Geo3DPruner, a geometry-guided 3D visual token pruning framework that effectively reduces redundant visual tokens while preserving multi-view consistency and maintaining the structural integrity of 3D scenes.
We first present the preliminaries of scene-level VideoLMs for 3D scene understanding in~\cref{Preliminaries}.
Subsequently, we describe the two stages of Geo3DPruner: Intra-Voxel View Consistency Pruning~(\cref{intra-voxel}), which exploits cross-view geometric consistency to identify and retain the most representative tokens within each voxel; and Inter-Voxel Spatial Diversity Pruning~(\cref{inter-voxel}), which further eliminates redundancy by selecting a geometrically diverse subset of voxels, thereby ensuring comprehensive coverage of the 3D scene.

\subsection{Preliminaries}
\label{Preliminaries}
In this study, we adopt Video-3D LLM~\cite{Video3DLLM} as the base model to demonstrate the effectiveness of our inference acceleration framework. 
Video-3D LLM~\cite{Video3DLLM} consists of two main components: a 2D visual encoder~(\ie, SigLIP~\cite{SigLIP}) and a LLM~(\ie, Qwen2~\cite{Qwen2}).
Since raw videos typically contain hundreds of frames and the LLM can only process a limited number of tokens due to GPU memory constraints, a uniform frame sampling strategy is applied to generate a manageable sequence~(\eg, 16 or 32 frames).

Given a sequence of RGB frames $\{\mathbf{I}_s \in \mathbb{R}^{H \times W \times 3}\}_{s=1}^{S}$, the visual encoder partitions each frame into patches of size $p$ and produces a corresponding grid of visual embeddings $\{\mathbf{E}_s \in \mathbb{R}^{H_p \times W_p \times d}\}_{s=1}^{S}$, where $H_p = \lfloor H / p \rfloor$, $W_p = \lfloor W / p \rfloor$, and $d$ denotes the feature dimension.
To construct a position-aware video representation, Video-3D LLM~\cite{Video3DLLM} transforms depth information into 3D coordinates and integrates these with the visual embeddings using 3D positional encodings. In our framework, these handcrafted positional encodings are replaced with geometry features extracted from 3D visual geometry models~(\ie, VGGT~\cite{VGGT}) following VG LLM~\cite{VGLLM}, which provide scene-consistent spatial cues that facilitate the subsequent token pruning process.
Let $\{\mathbf{G}_s \in \mathbb{R}^{H_p \times W_p \times d}\}_{s=1}^{S}$ denote the geometry features extracted from the $S$ input views, and let $\mathbf{A} \in \mathbb{R}^{N \times N}$ represent the global attention map produced by the cross-frame attention layers, where $N = S \times H_p \times W_p$.
We construct geometry-augmented visual features as $\{\mathbf{F}_s = \mathbf{E}_s + \mathbf{G}_s\}_{s=1}^{S}$ and feed them into the following modules. 


Despite its effectiveness, Video-3D LLM~\cite{Video3DLLM} incurs a substantial computational burden, as each frame generates hundreds of visual tokens, resulting in thousands of input tokens even after uniform frame sampling. This token explosion motivates our proposed Geo3DPruner framework, which efficiently eliminates redundant tokens while preserving multi-view consistency and maintaining the overall structural integrity of the 3D scene, as shown in~\cref{fig:framework}.

\subsection{Intra-Voxel View Consistency Pruning}
\label{intra-voxel}

\begin{figure}[t]
  \centering
  \includegraphics[width=1.0\linewidth]{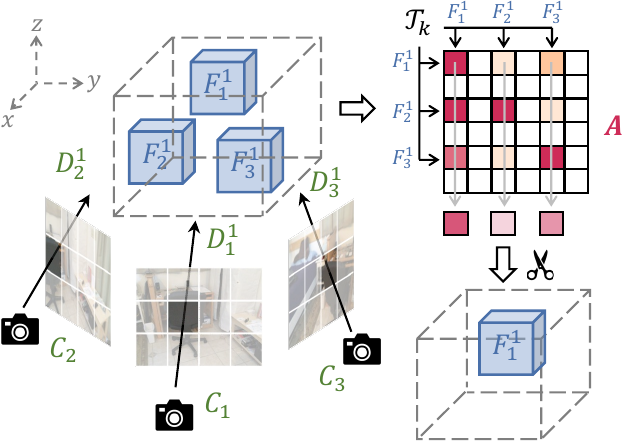}
  \caption{
  \textbf{Illustration of intra-voxel view consistency pruning.} Within each voxel, multi-view features from different frames are evaluated based on attention scores to identify the most representative tokens.
  }
  \vspace{-5mm}
  \label{fig:stage1}
\end{figure}

Video sequences can be naturally interpreted as multi-view observations of a 3D scene, where each frame provides a partial yet geometrically correlated view of the underlying structure. As a result, features corresponding to the same 3D location often recur across multiple viewpoints, leading to substantial redundancy.
To address this challenge, we propose Intra-Voxel View Consistency Pruning~(VCP), which explicitly models cross-view geometric correspondences and assesses the contribution of multi-view tokens within each voxel. 
As shown in~\cref{fig:stage1}, this mechanism ensures that only the most informative tokens are retained, effectively selecting observations from different views.

Within the intra-voxel pruning module, geometry features are further used to predict essential 3D attributes through task-specific heads, including camera parameters $\{\mathbf{C}_s \in \mathbb{R}^{9}\}_{s=1}^{S}$ and depth maps $\{\mathbf{D}_s \in \mathbb{R}^{H_p \times W_p}\}_{s=1}^{S}$. 
The camera parametes $\mathbf{C}_s$ can be decoded into camera extrinsics $\{\mathbf{R}_s \in \mathbb{R}^{3 \times 3}, \mathbf{T}_s \in \mathbb{R}^{3 \times 1}\}_{s=1}^{S}$ and intrinsics $\{\mathbf{Y}_s \in \mathbb{R}^{3 \times 3}\}_{s=1}^{S}$.
Given a token located at pixel coordinates $(u_s, v_s)$ in view $s$, its 3D position in the world coordinate system is computed via inverse camera projection.
Using a predefined voxel size $\delta$~(\eg, 0.1m), each token is assigned to a voxel.

Let $\mathcal{T}_k$ denote the set of token indices belonging to voxel $k$, with $|\mathcal{T}_k| = N_k$. 
To quantify cross-view geometric consistency, we extract the voxel-local attention submatrix from the global attention map: 
\begin{gather}
\mathbf{A}_k = \mathbf{A}[\mathcal{T}_k, \mathcal{T}_k] \in \mathbb{R}^{N_k \times N_k},
\end{gather}
where $\mathbf{A} \in \mathbb{R}^{N \times N}$ is the global attention map produced by the geometry encoder. 
The contribution score of token $i \in \mathcal{T}_k$ is then computed as the averaged incoming attention:
\begin{gather}
a_i = \frac{1}{N_k} \sum_{j \in \mathcal{T}_k} \mathbf{A}_k[j, i].
\end{gather}
This score evaluates how token $i$ is relative to all other multi-view features projected into the same voxel.
Finally, view consistency pruning retains only the top-$\alpha$ fraction of tokens in each voxel, suppressing redundant observations of identical 3D locations while preserving the most geometrically consistent and informative ones.

\subsection{Inter-Voxel Spatial Diversity Pruning}
\label{inter-voxel}

\begin{table*}[t]
  \caption{
  \textbf{Performance comparison with previous visual token pruning methods when the video sequence length is set to 16.}
  \textit{Avg.} represents the average percentage of performance maintained at the corresponding reduction ratio across five benchmarks and nine metrics.
  $\dagger$: We introduce an additional 3D encoder to replace the original 3D positional embeddings following VG LLM~\cite{VGLLM}.
  }
  \label{tab:frame16}
  \centering
  \setlength{\tabcolsep}{4pt}
  \begin{tabular}{c|cc|cc|cc|cc|c|c}
    \thickhline
    \multirow{2}{*}{Method} & \multicolumn{2}{c|}{ScanRefer} & \multicolumn{2}{c|}{Multi3DRefer} & \multicolumn{2}{c|}{Scan2Cap} & \multicolumn{2}{c|}{ScanQA} & SQA3D & \multirow{2}{*}{Avg.} \\
     & Acc@0.25 & Acc@0.5 & F1@0.25 & F1@0.5 & B-4@0.5 & C@0.5 & CIDEr & EM & EM & \\
    \hline \hline
    \rowcolor{title_gray}
    \multicolumn{11}{c}{\textit{16 Frames, All 3136 Tokens}} \\
    \hline
    Video-3D LLM$\dagger$~\cite{Video3DLLM} & 58.7 & 52.3 & 57.6 & 52.4 & 41.6 & 85.3 & 102.6 & 29.7 & 59.3 & 100\% \\
    \hline
    \rowcolor{title_gray}
    \multicolumn{11}{c}{\textit{16 Frames, Retain 1280 Tokens} \textcolor{blue}{($\downarrow$60\%)}} \\
    \hline
    FastV~\cite{FastV} & 55.8 & 49.6 & 54.4 & 49.6 & 39.2 & 73.1 & 99.6 & 29.0 & 57.6 & 94.5\% \\
    VisPruner~\cite{VisPruner} & 55.9 & 49.7 & 55.2 & 50.3 & 39.5 & 73.9 & 101.4 & 29.6 & 58.9 & 95.7\% \\
    Geo3DPruner~(Ours) & 58.4 & 52.0 & 56.7 & 51.7 & 41.6 & 85.1 & 99.2 & 29.1 & 59.3 & \textbf{98.9\%} \\
    \hline
    \rowcolor{title_gray}
    \multicolumn{11}{c}{\textit{16 Frames, Retain 640 Tokens} \textcolor{blue}{($\downarrow$80\%)}} \\
    \hline
    FastV~\cite{FastV} & 53.4 & 47.7 & 51.1 & 46.7 & 36.7 & 62.7 & 93.2 & 27.4 & 55.6 & 88.7\% \\
    VisPruner~\cite{VisPruner} & 54.3 & 48.3 & 52.8 & 48.2 & 37.7 & 66.6 & 97.8 & 28.5 & 56.7 & 91.6\% \\
    Geo3DPruner~(Ours) & 57.2 & 51.1 & 54.0 & 49.5 & 40.9 & 82.9 & 96.1 & 28.0 & 58.1 & \textbf{96.1\%} \\
    \hline
    \rowcolor{title_gray}
    \multicolumn{11}{c}{\textit{16 Frames, Retain 320 Tokens} \textcolor{blue}{($\downarrow$90\%)}} \\
    \hline
    FastV~\cite{FastV} & 51.2 & 45.8 & 44.0 & 40.5 & 34.5 & 53.0 & 84.7 & 24.6 & 53.5 & 81.0\% \\
    VisPruner~\cite{VisPruner} & 52.4 & 46.8 & 47.9 & 43.9 & 35.4 & 57.3 & 88.7 & 26.0 & 54.6 & 84.9\% \\
    Geo3DPruner~(Ours) & 55.2 & 49.4 & 51.1 & 46.9 & 40.6 & 80.3 & 91.3 & 26.0 & 55.7 & \textbf{92.1\%} \\
    \hline
  \end{tabular}
\end{table*}

\begin{table*}[t]
  \caption{
  \textbf{Performance comparison with previous visual token pruning methods when the video sequence length is set to 32.}
  \textit{Avg.} represents the average percentage of performance maintained at the corresponding reduction ratio across five benchmarks and nine metrics.
  $\dagger$: We introduce an additional 3D encoder to replace the original 3D positional embeddings following VG LLM~\cite{VGLLM}.
  }
  \label{tab:frame32}
  \centering
  \setlength{\tabcolsep}{4pt}
  \begin{tabular}{c|cc|cc|cc|cc|c|c}
    \thickhline
    \multirow{2}{*}{Method} & \multicolumn{2}{c|}{ScanRefer} & \multicolumn{2}{c|}{Multi3DRefer} & \multicolumn{2}{c|}{Scan2Cap} & \multicolumn{2}{c|}{ScanQA} & SQA3D & \multirow{2}{*}{Avg.} \\
     & Acc@0.25 & Acc@0.5 & F1@0.25 & F1@0.5 & B-4@0.5 & C@0.5 & CIDEr & EM & EM & \\
    \hline \hline
    \rowcolor{title_gray}
    \multicolumn{11}{c}{\textit{32 Frames, All 6272 Tokens}} \\
    \hline
    Video-3D LLM$\dagger$~\cite{Video3DLLM} & 62.0 & 55.1 & 60.1 & 54.6 & 42.6 & 89.0 & 104.3 & 29.8 & 60.3 & 100\% \\
    \hline
    \rowcolor{title_gray}
    \multicolumn{11}{c}{\textit{32 Frames, Retain 2560 Tokens} \textcolor{blue}{($\downarrow$60\%)}} \\
    \hline
    FastV~\cite{FastV} & 59.9 & 53.3 & 57.5 & 52.3 & 39.8 & 77.7 & 100.6 & 29.0 & 58.7 & 95.2\% \\
    VisPruner~\cite{VisPruner} & 60.2 & 53.5 & 58.5 & 53.1 & 39.7 & 77.4 & 101.5 & 29.1 & 58.9 & 95.7\% \\
    Geo3DPruner~(Ours) & 61.3 & 54.6 & 59.0 & 53.7 & 42.4 & 87.3 & 101.3 & 29.2 & 59.6 & \textbf{98.5\%} \\
    \hline
    \rowcolor{title_gray}
    \multicolumn{11}{c}{\textit{32 Frames, Retain 1280 Tokens} \textcolor{blue}{($\downarrow$80\%)}} \\
    \hline
    FastV~\cite{FastV} & 57.6 & 51.2 & 53.6 & 48.9 & 36.9 & 63.6 & 93.3 & 26.7 & 56.6 & 88.4\% \\
    VisPruner~\cite{VisPruner} & 58.6 & 52.2 & 56.1 & 51.1 & 37.9 & 68.2 & 97.4 & 28.0 & 57.4 & 91.6\% \\
    Geo3DPruner~(Ours) & 60.1 & 53.5 & 57.5 & 52.5 & 42.1 & 85.7 & 97.9 & 28.3 & 58.3 & \textbf{96.3\%} \\
    \hline
    \rowcolor{title_gray}
    \multicolumn{11}{c}{\textit{32 Frames, Retain 640 Tokens} \textcolor{blue}{($\downarrow$90\%)}} \\
    \hline
    FastV~\cite{FastV} & 55.7 & 49.6 & 47.3 & 43.4 & 34.9 & 57.0 & 86.9 & 24.8 & 54.7 & 82.4\% \\
    VisPruner~\cite{VisPruner} & 56.2 & 50.1 & 50.6 & 46.4 & 35.8 & 57.3 & 90.4 & 26.0 & 54.6 & 84.8\% \\
    Geo3DPruner~(Ours) & 58.0 & 51.8 & 53.9 & 49.3 & 41.3 & 82.3 & 92.1 & 26.8 & 55.8 & \textbf{92.0\%} \\
    \hline
  \end{tabular}
\end{table*}

After pruning redundant tokens within each voxel, we formulate the subsequent voxel-level pruning as a subset selection problem, aiming to identify an optimal subset of voxels that maximizes scene representation under a fixed budget.
Existing training-free pruning strategies typically operate on per-frame features without a global 3D modeling mechanism, causing them to retain visually salient tokens within each frame. 
However, in object-centric 3D scenes with high instance diversity, such methods often introduce strong intra-object bias, whereby the retained tokens concentrate on the same object across multiple views, hindering effective reconstruction of the complete 3D environment. As a result, model performance deteriorates sharply as the token budget decreases.
To address this limitation, we propose Inter-Voxel Spatial Diversity Pruning~(SDP), an iterative heuristic designed to optimize this selection. At each iteration, SDP adaptively selects a small number of voxels with the highest global importance and then recalculates the relevance scores among the remaining voxel candidates. This process explicitly suppresses redundant attention concentration within the same object while progressively enhancing instance-level diversity across the entire scene.

Specifically, given two voxels $k$ and $l$ with their corresponding token sets $\mathcal{T}_k$ and $\mathcal{T}_l$, we extract the cross-voxel attention submatrix from the global attention map as:
\begin{gather}
\mathbf{A}_{k \rightarrow l} = \mathbf{A}[\mathcal{T}_k, \mathcal{T}_l] \in \mathbb{R}^{|\mathcal{T}_k| \times |\mathcal{T}_l|}.
\end{gather}
To quantify the extent to which voxel $k$ attends to voxel $l$, we first sum over rows and then average over columns:
\begin{gather}
a_{k \rightarrow l} = \frac{1}{|\mathcal{T}_k|} \sum_{j=1}^{|\mathcal{T}_k|} \left( \sum_{i=1}^{|\mathcal{T}_l|} \mathbf{A}_{k \rightarrow l}(j, i) \right).
\end{gather}
The global attention received by voxel $l$ from all other voxels is then computed as:
\begin{gather}
a_l = \sum_{k} a_{k \rightarrow l}.
\end{gather}
A high value of $a_l$ indicates that voxel $l$ receives a large amount of attention from across the scene.
However, performing a one-shot pruning to the target pruning ratio solely based on $a_l$ leads to severe redundancy, as voxels belonging to the same object typically exhibit mutually reinforcing attention and would all be retained.
To alleviate this intra-object bias, we adopt an iterative selection cycle.

Let $\mathcal{V}$ denote the set of voxel candidates and $\mathcal{W}$ the set of selected voxels, initialized as $\mathcal{W} = \varnothing$. 
The spatial diversity pruning iteratively selects voxels based on the voxel-wise attention while suppressing redundancy within the same instance. 
In each iteration, attention scores are computed for all remaining candidate voxels in $\mathcal{V} \setminus \mathcal{W}$ to assess their relative importance. 
The top-$K$ salient voxels $\Delta \mathcal{W}$ are then selected and incorporated into the current set $\mathcal{W} \leftarrow \mathcal{W} \cup \Delta \mathcal{W}$. 
Subsequently, attention among the remaining candidates is recomputed, considering only unselected voxels. 
This iterative procedure continues until the total number of tokens in the selected voxels meets the target budget. 

By repeatedly suppressing voxel redundancy through global attention recomputation, our spatial diversity pruning prevents high-attention voxels belonging to the same instance from dominating the token allocation. Consequently, the final set of selected voxels covers the most informative and spatially diverse regions, preserving the essential 3D structure of the scene under strict token constraints.


%% file: sec/4_experiments.tex
\section{Experiments}
\subsection{Datasets and Metrics}
ScanNet~\cite{ScanNet} is a large-scale RGB-D video dataset consisting of 1,513 indoor scenes that cover diverse 3D environments, such as offices, apartments, and hallways.
Each scene is annotated with 3D reconstructions, camera poses, depth maps, and object-level semantic segmentation, making ScanNet~\cite{ScanNet} a widely used benchmark for 3D scene understanding. 
We conduct experiments on 3D visual grounding, 3D dense captioning, and 3D question answering. For all tasks, the benchmarks are derived from ScanNet~\cite{ScanNet}, with task-pecific annotations. 
Although depth maps are provided, we still employ VGGT~\cite{VGGT} to extract geometry-aware features, as raw depth alone does not encode high-level geometric priors or cross-view structural consistency required by our geometry-guided pruning framework.

\noindent \textbf{3D visual grounding.}
For 3D visual grounding, we evaluate our method on ScanRefer~\cite{ScanRefer} and Multi3DRefer~\cite{Multi3DRefer}, which involve localizing one or more objects using natural language descriptions.
For ScanRefer, we report the thresholded accuracy based on Intersection over Union~(IoU) thresholds of 0.25 and 0.5.
For Multi3DRefer, we evaluate the F1 score at IoU thresholds of 0.25 and 0.5.

\noindent \textbf{3D dense captioning.}
For 3D dense captioning, we use the Scan2Cap~\cite{Scan2Cap} benchmark, which predicts all the bounding boxes in the scene along with descriptions of the objects.
To evaluate the descriptions, we combine traditional image captioning metrics, such as BLEU-4~\cite{BLEU} and CIDEr~\cite{CIDEr}, with an IoU threshold of 0.5 between the predicted and ground-truth bounding boxes.

\noindent \textbf{3D question answering.}
For 3D question answering, we use ScanQA~\cite{ScanQA} for spatial reasoning and SQA3D~\cite{SQA3D} for situated reasoning.
For ScanQA~\cite{ScanQA}, we use CIDEr~\cite{CIDEr} and exact match~(EM) as evaluation metrics.
For SQA3D~\cite{SQA3D}, we only adopt exact match~(EM) as the metric.

\subsection{Implementation Details}
We adopt Video-3D LLM~\cite{Video3DLLM} as our baseline, which is built upon LLaVA-Video 7B~\cite{LLaVA-Video}. The 2D visual encoder is SigLIP~\cite{SigLIP}, and the large language model is Qwen2-7B~\cite{Qwen2}.
To incorporate 3D geometric information, following VG LLM~\cite{VGLLM}, we introduce VGGT-1B~\cite{VGGT} as an additional 3D encoder to replace the original hand-crafted 3D positional embeddings.
All training settings of Video-3D LLM~\cite{Video3DLLM} remain consistent with the default configuration in its official open-source codebase, including the optimizer, learning rate, and training schedule.
The parameters of VGGT~\cite{VGGT} are frozen during training to leverage pre-trained 3D feature representations without additional optimization.
For each scene, video frames are uniformly sampled from the original sequence. The frames are preprocessed with resizing and center cropping to a resolution of $384\times384$.
During inference, Geo3DPruner is applied to remove redundant visual tokens. The voxel size $\delta$ is set to 0.1m. In the intra-voxel pruning module, the pruning ratio $\alpha$ is set to 50\%, retaining only the most representative tokens within each voxel. In the inter-voxel pruning stage, the iterative attention-based selection selects $K=8$ top voxels at each iteration until the token budget is met.

\begin{table}[t]
  \caption{
  \textbf{Impact of individual pruning stages.}
  \textit{VCP} denotes the Intra-Voxel View Consistency Pruning stage and \textit{SDP} denotes the Inter-Voxel Spatial Diversity Pruning stage.
  }
  \label{tab:stage}
  \centering
  \setlength{\tabcolsep}{4pt}
  \begin{tabular}{cc|ccc|c}
    \thickhline
    VCP & SDP & ScanRefer & Scan2Cap & SQA3D & Avg. \\
    \hline \hline
    $\checkmark$ & & 52.5 & 76.7 & 52.7 & 89.4\% \\
    & $\checkmark$ & 53.3 & 77.7 & 54.6 & 91.3\% \\
    $\checkmark$ & $\checkmark$ & 55.2 & 80.3 & 55.7 & \textbf{94.0\%} \\
    \hline
  \end{tabular}
\end{table}


\begin{table}[t]
  \caption{
  \textbf{Comparison of different voxel selection strategies.}
  \textit{Rand.} randomly samples a subset of voxels, while \textit{Unif.} samples voxels uniformly across spatial regions.
  }
  \label{tab:voxel}
  \centering
  \setlength{\tabcolsep}{4pt}
  \begin{tabular}{c|ccc|c}
    \thickhline
    Method & ScanRefer & Scan2Cap & SQA3D & Avg. \\
    \hline \hline
    Rand. & 53.0 & 61.4 & 55.4 & 85.2\% \\
    Unif. & 53.2 & 62.0 & 55.0 & 85.4\% \\
    SDP & 55.2 & 80.3 & 55.7 & \textbf{94.0\%} \\
    \hline
  \end{tabular}
\end{table}

\begin{table}[t]
  \caption{
  \textbf{Strategy for computing global cross-frame relevance.}
  \textit{Attn.} denotes the attention-based method and \textit{Sim.} denotes the similarity-based method.
  }
  \label{tab:attn}
  \centering
  \setlength{\tabcolsep}{4pt}
  \begin{tabular}{c|ccc|c}
    \thickhline
    Method & ScanRefer & Scan2Cap & SQA3D & Avg. \\
    \hline \hline
    Sim. & 52.9 & 76.4 & 53.3 & 89.9\% \\
    Attn. & 55.2 & 80.3 & 55.7 & \textbf{94.0\%} \\
    \hline
  \end{tabular}
\end{table}

\subsection{Main results}
We conduct a comprehensive evaluation of Geo3DPruner by comparing it with existing visual token pruning methods under varying video lengths and pruning ratios. 
\cref{tab:frame16} reports the performance of different pruning strategies when retaining 1,280, 640, and 320 tokens, corresponding to pruning ratios of 60\%, 80\%, and 90\%, respectively.
At a moderate pruning ratio of 60\%, Geo3DPruner preserves nearly all of the original performance, achieving 98.9\% of the baseline accuracy. 
When the pruning ratio increases to 80\%, our method still maintains strong performance, with only a 3.9\% drop relative to the unpruned model.
Even under extreme pruning, retaining only 10\% of the tokens, Geo3DPruner achieves 92.1\% of the baseline accuracy, substantially outperforming both text-guided method FastV~\cite{FastV} and vision-guided method VisPruner~\cite{VisPruner}. 
These results highlight the effectiveness of our geometry-guided design, which \textit{leverages cross-frame global attention to jointly enforce view-consistency and spatial diversity}, thereby preserving the structural integrity of the 3D scene even under aggressive token reduction.

We further investigate the scalability of Geo3DPruner on longer videos consisting of 32 frames, doubling the number of visual tokens to 6,272. 
As reported in~\cref{tab:frame32}, Geo3DPruner maintains high performance even at a 90\% pruning ratio, retaining only 640 tokens for the entire video while preserving 92.0\% of the original performance. In contrast, FastV~\cite{FastV} and VisPruner~\cite{VisPruner} achieve only 82.4\% and 84.8\%, respectively, under the same pruning conditions.

\begin{figure*}[t]
  \centering
  \includegraphics[width=1.0\linewidth]{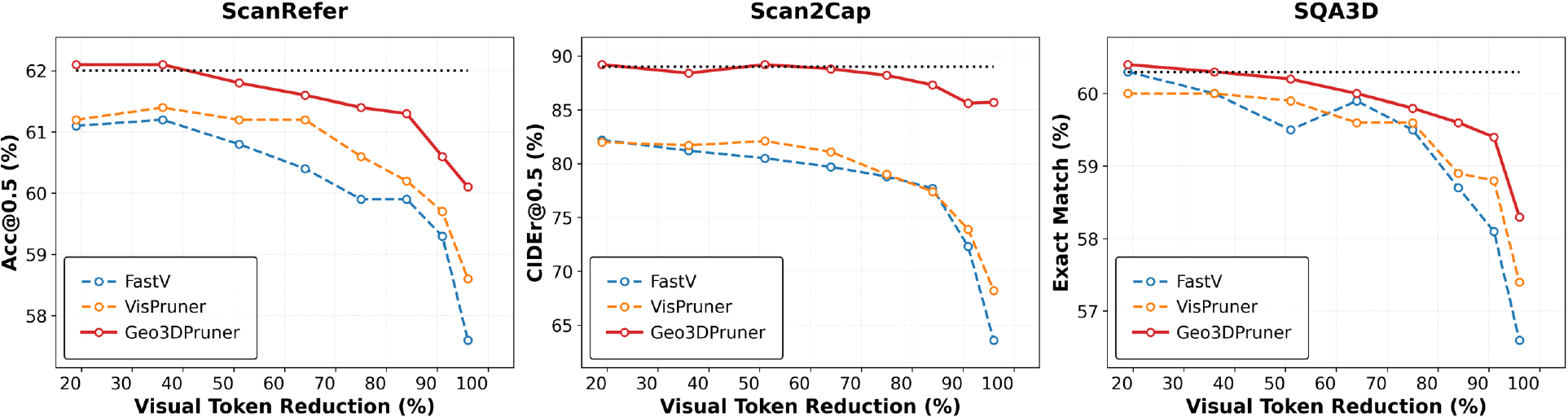}
  \caption{
  \textbf{Performance-efficiency trade-off curves for 32-frame videos using different visual token pruning methods.} The x-axis denotes the reduction ratio in the number of visual tokens, and the y-axis shows the corresponding performance across benchmarks. The black dotted line represents the unpruned base model.
  }
  \vspace{-5mm}
  \label{fig:flops}
\end{figure*}

\subsection{Ablation Studies}
We conduct comprehensive ablation studies to analyze the contributions of individual components in Geo3DPruner. All experiments use 16-frame videos with a 90\% pruning ratio.
We report results on three representative benchmarks: Acc@0.5 on ScanRefer~\cite{ScanRefer} for 3D visual grounding, CIDEr@0.5 on Scan2Cap~\cite{Scan2Cap} for 3D dense captioning, and EM on SQA3D~\cite{SQA3D} for 3D question answering. 

\noindent \textbf{Impact of individual pruning stages.}
To evaluate the effectiveness of the two-stage pruning design, we consider three variants, applying only intra-voxel VCP, only inter-voxel SDP, and the two-stage pipeline.
As shown in~\cref{tab:stage}, using VCP alone ensures that tokens are retained across most spatial locations, but the features within each voxel become overly sparse, limiting the overall performance to 89.4\% of the baseline.
In contrast, relying solely on SDP without first removing intra-voxel redundancy causes excessive voxel pruning, reducing spatial coverage and resulting in an 8.7\% drop relative to the unpruned model.
The integration of both VCP and SDP consistently delivers the highest performance across all benchmarks, highlighting the complementary roles of intra-voxel and inter-voxel pruning in removing redundant multi-view features while preserving the structural completeness of the 3D scene.


\noindent \textbf{Comparison of different voxel selection strategies.}
We formulate voxel-level pruning as a subset selection problem and propose an iterative heuristic to approximate its solution. To validate the effectiveness of our strategy, we compare it with two naive baselines. 
The random selection method randomly samples a subset of voxels from all candidates, while the uniform selection method uniformly samples voxels across the spatial domain to ensure coarse spatial coverage. 
The results are reported in~\cref{tab:voxel}. Compared with these naive subset selection strategies, our iterative heuristic consistently achieves substantial performance improvements, highlighting the importance of diversity-aware selection in solving this optimization problem.

\noindent \textbf{Strategy for computing global cross-frame relevance.}
We further investigate two alternative strategies for modeling global cross-frame relevance, as summarized in~\cref{tab:attn}.
The attention-based method utilizes the attention maps produced by the cross-frame attention layer within the geometry encoder, whereas the similarity-based method computes pairwise cosine similarity between geometry features extracted from different frames.
Across all benchmarks, the attention-based method consistently outperforms its similarity-based counterpart.
This improvement stems from the adaptive capability of the attention mechanism, which dynamically emphasizes cross-view features that are geometrically consistent while suppressing noisy correlations.
In contrast, the similarity-based method relies solely on feature-level similarity, making it less effective in capturing the complex global dependencies across frames.

\begin{figure}[t]
  \centering
  \includegraphics[width=1.0\linewidth]{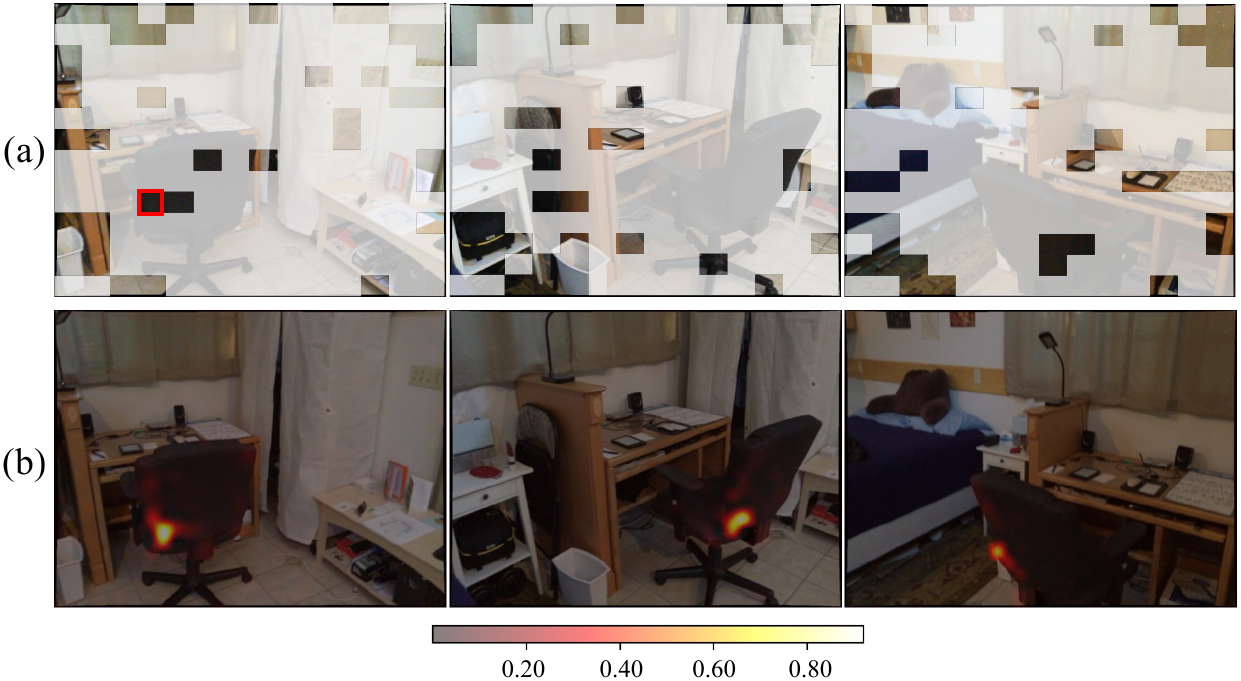}
  \caption{
  (a) \textbf{Retained tokens in each frame} show diverse spatial coverage across different objects, preserving overall scene completeness.
  (b) \textbf{Global attention maps} from the red-box patch in the first frame primarily focus on geometry-aligned and instance-related regions in other frames.
  }
  \vspace{-5mm}
  \label{fig:vis}
\end{figure}

\subsection{Efficiency analysis}
The performance-efficiency curves for 32-frame videos are presented in~\cref{fig:flops}.
Across all pruning ratios, Geo3DPruner consistently outperforms FastV and VisPruner, demonstrating the effectiveness of our cross-frame relevance modeling.
This advantage is particularly pronounced on object-centric benchmarks~(\eg, Scan2Cap), where preserving spatial diversity is critical for maintaining fine-grained scene completeness. 
Remarkably, when pruning a small fraction of tokens~(\ie, 20\%), the performance of Geo3DPruner slightly exceeds that of the uncompressed baseline. 
This observation indicates that our method indeed removes redundant or noisy visual tokens and produces a more compact representation of the 3D scene.

\subsection{Visualization}
As shown in~\cref{fig:vis}~(a), we visualize the retained tokens in each frame. The preserved tokens exhibit diverse spatial coverage across different objects in the scene, confirming that our pruning strategy effectively maintains spatial completeness.
Furthermore, for the retained feature within the red-box patch in the first frame, we visualize the global attention maps from this patch to all patches across all frames in~\cref{fig:vis}~(b). This patch primarily attends to geometry-aligned and instance-related regions in other frames, indicating that the geometry-guided global attention effectively models cross-frame correspondences.
Meanwhile, the corresponding patches in other frames are pruned, demonstrating that our method successfully removes multi-view redundancy while retaining the most informative representations.

%% file: sec/5_conclusion.tex
\section{Conclusion}
We propose Geo3DPruner, a geometry-guided framework for 3D visual token pruning in video-language models. It models global cross-frame relevance using geometry-aware attention and employs a two-stage strategy with intra-voxel view-consistent selection and inter-voxel spatially diverse pruning. Our method reduces token redundancy while preserving the 3D scene structure. Experiments on multiple benchmarks show that Geo3DPruner improves inference efficiency while maintaining strong task performance.

%% file: sec/X_suppl.tex
\clearpage
\appendix
\setcounter{page}{1}
\maketitlesupplementary

\section{Dataset Details}
\noindent \textbf{ScanRefer~\cite{ScanRefer}.}
ScanRefer~\cite{ScanRefer} is a large-scale 3D visual grounding dataset built on ScanNet~\cite{ScanNet}, providing natural-language descriptions of indoor scenes to support the task of localizing target objects in 3D environments based on free-form textual queries.
The dataset contains 51,583 descriptions from 800 ScanNet~\cite{ScanNet} scenes, covering 11,046 annotated objects, with an average of 13.81 objects and 64.48 descriptions per scene. The annotations are diverse and semantically rich, spanning more than 250 indoor object categories and frequently incorporating spatial language~(98.7\%), color~(74.7\%), and shape terms~(64.9\%).
Following the official ScanNet~\cite{ScanNet} split, the dataset is divided into 36,665 training, 9,508 validation, and 5,410 test samples, with no scene overlap across splits.
Evaluation is performed using thresholded Intersection over Union~(IoU) accuracy, where a prediction is considered correct if its IoU with the ground-truth bounding box exceeds 0.25 or 0.5, offering a reliable measure of grounding quality.

\noindent \textbf{Multi3DRefer~\cite{Multi3DRefer}.}
Multi3DRefer~\cite{Multi3DRefer} extends ScanRefer~\cite{ScanRefer} to support grounding tasks involving a flexible number of target objects in real-world 3D scenes.
The dataset includes 61,926 descriptions covering 11,609 objects, of which 51,583 are inherited from ScanRefer~\cite{ScanRefer}, 6,688 describe zero-target situations, and 13,178 involve multiple target objects. Scenes with multiple targets are typically offices or meeting rooms containing many chairs and tables.
To evaluate grounding performance under variable target counts, Multi3DRefer~\cite{Multi3DRefer} employs F1 scores at IoU thresholds of 0.25 and 0.5. During evaluation, per-pair IoUs between predicted and ground-truth boxes are computed, and the Hungarian algorithm is applied to obtain an optimal one-to-one matching. Pairs with IoUs exceeding the threshold are considered true positives. For zero-target cases, recall is set to 1, while precision is set to 1 if no predictions are made, and 0 otherwise.

\noindent \textbf{Scan2Cap~\cite{Scan2Cap}.}
Scan2Cap~\cite{Scan2Cap} is a 3D dense captioning dataset built upon ScanRefer~\cite{ScanRefer}, targeting the joint task of object detection and natural-language description generation in 3D scenes.
Each description provides information about the appearance of the object~(\eg, ``a black wooden chair'') as well as its spatial relationships with surrounding objects~(\eg, ``the chair is placed at the end of the long dining table, just before the TV mounted on the wall'').
Following the ScanRefer~\cite{ScanRefer} split, the dataset contains 36,665 training samples and 9,508 validation samples.
To jointly evaluate the quality of generated descriptions and the accuracy of detected bounding boxes, descriptions are assessed using standard image captioning metrics such as CIDEr~\cite{CIDEr} and BLEU-4~\cite{BLEU}, combined with IoU scores between predicted and target bounding boxes.

\noindent \textbf{ScanQA~\cite{ScanQA}.}
ScanQA~\cite{ScanQA} is a large-scale 3D question answering dataset built on ScanNet~\cite{ScanNet}, designed to evaluate spatial understanding in 3D indoor scenes.
In this task, models receive a full 3D reconstructed scene and answer free-form textual questions regarding objects and their spatial relations.
The dataset contains 41,363 questions and 58,191 answers, including 32,337 unique questions and 16,999 unique answers. To account for variability in free-form responses, at least two ground-truth answers are provided for each validation and test question.
Evaluation uses exact match~(EM@$K$), which measures the proportion of samples for which any of the top-$K$ predictions exactly matches one of the ground-truth answers. In addition, standard image captioning metrics, such as CIDEr~\cite{CIDEr}, are applied to capture semantic similarity and enable robust evaluation for questions with multiple valid answer expressions.

\noindent \textbf{SQA3D~\cite{SQA3D}.}
SQA3D~\cite{SQA3D} is a large-scale benchmark aimed at evaluating situated scene understanding in 3D environments.
Given a 3D scene, the benchmark requires an agent to infer its situated context~(\eg, position and orientation) from a textual description, and subsequently reason about the surrounding environment to answer questions grounded in that situation.
Built on 650 ScanNet~\cite{ScanNet} scenes, SQA3D~\cite{SQA3D} comprises 6.8k unique situations, 20.4k descriptive texts, and 33.4k diverse reasoning questions, covering spatial relations, navigation, common sense reasoning, and multi-hop reasoning.
Following the ScanNet~\cite{ScanNet} data split, the dataset is divided into training, validation, and test sets.
Evaluation relies on exact match~(EM) accuracy as the primary metric, which is found to be sufficiently discriminative for comparing model performance.

\begin{table*}[t]
  \caption{
  \textbf{Performance comparison with previous visual token pruning methods when the video sequence length is set to 8.}
  \textit{Avg.} represents the average percentage of performance maintained at the corresponding reduction ratio across five benchmarks and nine metrics.
  $\dagger$: Following VG LLM~\cite{VGLLM}, we introduce an additional 3D encoder to replace the original 3D positional embeddings.
  }
  \label{tab:frame8}
  \centering
  \setlength{\tabcolsep}{4pt}
  \begin{tabular}{c|cc|cc|cc|cc|c|c}
    \thickhline
    \multirow{2}{*}{Method} & \multicolumn{2}{c|}{ScanRefer} & \multicolumn{2}{c|}{Multi3DRefer} & \multicolumn{2}{c|}{Scan2Cap} & \multicolumn{2}{c|}{ScanQA} & SQA3D & \multirow{2}{*}{Avg.} \\
     & Acc@0.25 & Acc@0.5 & F1@0.25 & F1@0.5 & B-4@0.5 & C@0.5 & CIDEr & EM & EM & \\
    \hline \hline
    \rowcolor{title_gray}
    \multicolumn{11}{c}{\textit{8 Frames, All 1568 Tokens}} \\
    \hline
    Video-3D LLM$\dagger$~\cite{Video3DLLM} & 52.1 & 46.3 & 51.5 & 47.0 & 39.9 & 76.0 & 99.2 & 28.0 & 58.0 & 100\% \\
    \hline
    \rowcolor{title_gray}
    \multicolumn{11}{c}{\textit{8 Frames, Retain 640 Tokens} \textcolor{blue}{($\downarrow$60\%)}} \\
    \hline
    FastV~\cite{FastV} & 48.7 & 43.1 & 48.1 & 43.9 & 35.4 & 62.7 & 93.0 & 26.7 & 56.8 & 92.4\% \\
    VisPruner~\cite{VisPruner} & 49.7 & 44.3 & 49.2 & 45.0 & 35.5 & 62.3 & 94.1 & 27.5 & 56.9 & 93.8\% \\
    Geo3DPruner~(Ours) & 50.9 & 45.4 & 50.0 & 45.7 & 39.7 & 74.4 & 95.2 & 27.2 & 57.2 & \textbf{97.7\%} \\
    \hline
    \rowcolor{title_gray}
    \multicolumn{11}{c}{\textit{8 Frames, Retain 320 Tokens} \textcolor{blue}{($\downarrow$80\%)}} \\
    \hline
    FastV~\cite{FastV} & 45.0 & 40.0 & 43.8 & 40.2 & 33.6 & 51.0 & 88.7 & 25.8 & 54.9 & 85.7\% \\
    VisPruner~\cite{VisPruner} & 46.8 & 41.6 & 46.5 & 42.5 & 33.9 & 54.3 & 89.7 & 26.1 & 55.1 & 88.4\% \\
    Geo3DPruner~(Ours) & 48.4 & 43.1 & 47.1 & 43.3 & 39.1 & 70.6 & 92.5 & 26.5 & 56.7 & \textbf{94.0\%} \\
    \hline
    \rowcolor{title_gray}
    \multicolumn{11}{c}{\textit{8 Frames, Retain 160 Tokens} \textcolor{blue}{($\downarrow$90\%)}} \\
    \hline
    FastV~\cite{FastV} & 41.5 & 36.7 & 37.7 & 34.6 & 32.4 & 45.1 & 83.4 & 23.9 & 53.5 & 78.7\% \\
    VisPruner~\cite{VisPruner} & 42.8 & 38.0 & 40.6 & 37.2 & 33.5 & 48.6 & 84.9 & 24.3 & 53.4 & 81.6\% \\
    Geo3DPruner~(Ours) & 44.8 & 40.2 & 42.0 & 38.9 & 39.2 & 70.2 & 86.6 & 25.1 & 54.4 & \textbf{88.7\%} \\
    \hline
  \end{tabular}
\end{table*}

\begin{table*}[t]
  \caption{
  \textbf{Comparison with state-of-the-art methods} on 3D scene understanding benchmarks.
  The video sequence lengths for Video-3D LLM~\cite{Video3DLLM} and VG LLM~\cite{VGLLM} are set to 32.
  $\dagger$: Following VG LLM~\cite{VGLLM}, we introduce an additional 3D encoder to replace the original 3D positional embeddings.
  }
  \label{tab:sota}
  \centering
  \setlength{\tabcolsep}{4pt}
  \begin{tabular}{c|cc|cc|cc|cc|c}
    \thickhline
    \multirow{2}{*}{Method} & \multicolumn{2}{c|}{ScanRefer} & \multicolumn{2}{c|}{Multi3DRefer} & \multicolumn{2}{c|}{Scan2Cap} & \multicolumn{2}{c|}{ScanQA} & SQA3D \\
    & Acc@0.25 & Acc@0.5 & F1@0.25 & F1@0.5 & B-4@0.5 & C@0.5 & CIDEr & EM & EM \\
    \hline \hline
    ScanRefer~\cite{ScanRefer} & 37.3 & 24.3 & - & - & - & - & - & - & - \\
    M3DRef-CLIP~\cite{Multi3DRefer} & 51.9 & 44.7 & 42.8 & - & 38.4 & - & - & - & - \\
    Scan2Cap~\cite{Scan2Cap} & - & - & - & - & 22.4 & 35.2 & - & - & - \\
    ScanQA~\cite{ScanQA} & - & - & - & - & - & - & 64.9 & 21.1 & 47.2 \\
    3D-LLM~(Flamingo)~\cite{3D-LLM} & 21.2 & - & - & - & - & - & 59.2 & 20.4 & - \\
    3D-LLM~(BLIP2-FlanT5)~\cite{3D-LLM} & 30.3 & - & - & - & - & - & 69.4 & 20.5 & - \\
    LL3DA~\cite{LL3DA} & - & - & - & - & 36.0 & 62.9 & 76.8 & - & - \\
    LEO~\cite{LEO} & - & - & - & - & 38.2 & 72.4 & 101.4 & 21.5 & 50.0 \\
    LLaVA-3D~\cite{LLaVA-3D} & 54.1 & 42.4 & - & - & 41.1 & 79.2 & 91.7 & 27.0 & 55.6 \\
    Video-3D LLM~\cite{Video3DLLM} & 58.1 & 51.7 & 58.0 & 52.7 & 41.3 & 83.8 & 102.1 & \textbf{30.1} & 58.6 \\
    VG LLM~\cite{VGLLM} & 57.6 & 50.9 & - & - & 41.5 & 80.0 & - & - & - \\
    \hline
    Video-3D LLM$\dagger$~\cite{Video3DLLM} & \textbf{62.0} & \textbf{55.1} & \textbf{60.1} & \textbf{54.6} & \textbf{42.6} & \textbf{89.0} & \textbf{104.3} & 29.8 & \textbf{60.3} \\
    + Geo3DPruner~(\textcolor{blue}{$\downarrow$60\%}) & 61.3 & 54.6 & 59.0 & 53.7 & 42.4 & 87.3 & 101.3 & 29.2 & 59.6 \\
    + Geo3DPruner~(\textcolor{blue}{$\downarrow$80\%}) & 60.1 & 53.5 & 57.5 & 52.5 & 42.1 & 85.7 & 97.9 & 28.3 & 58.3 \\
    + Geo3DPruner~(\textcolor{blue}{$\downarrow$90\%}) & 58.0 & 51.8 & 53.9 & 49.3 & 41.3 & 82.3 & 92.1 & 26.8 & 55.8 \\
    \hline
  \end{tabular}
\end{table*}

\section{More Quantitative Results}
\subsection{Comparison under the 8-Frame Video Setting}
We additionally evaluate Geo3DPruner on a shorter video setting with 8-frame inputs, where each scene contains only 1,568 visual tokens. This reduced video context provides a weaker representation of the 3D scene, making token pruning more challenging.
\cref{tab:frame8} reports the performance when retaining 640, 320, and 160 tokens, corresponding to pruning ratios of 60\%, 80\%, and 90\%, respectively.
Geo3DPruner preserves nearly all of the baseline performance at a pruning ratio of 60\%, achieving 97.7\% of the unpruned accuracy. Increasing the pruning ratio to 80\% results in only a slight performance degradation, demonstrating the robustness of our method even with limited visual information. With only 160 tokens retained under the extreme 90\% pruning setting, Geo3DPruner maintains 88.7\% of the baseline accuracy, substantially outperforming FastV~\cite{FastV} and VisPruner~\cite{VisPruner}, which experience significant drops under the same conditions.
These results indicate that even with limited 3D coverage, geometry-aware cross-frame correspondence enables effective token selection and robust scene understanding.

\subsection{Comparison with State-of-the-Art Methods}
We further evaluate the model pruned by Geo3DPruner against state-of-the-art methods, as summarized in~\cref{tab:sota}. By incorporating geometry features, Video-3D LLM$\dagger$~\cite{Video3DLLM} consistently outperforms existing methods across all benchmarks. After pruning 90\% of the visual tokens, Video-3D LLM$\dagger$~\cite{Video3DLLM} with Geo3DPruner maintains performance comparable to state-of-the-art methods while operating on a substantially reduced number of tokens.

\begin{table*}[t]
  \caption{
  \textbf{Comparison under a fixed token budget} of 1568 tokens. The baseline uses 8 input frames without pruning, while Geo3DPruner leverages more frames with corresponding pruning ratios to maintain the same number of tokens.
  \textit{Avg.} represents the average percentage of performance maintained at the corresponding reduction ratio across five benchmarks and nine metrics.
  $\dagger$: Following VG LLM~\cite{VGLLM}, we introduce an additional 3D encoder to replace the original 3D positional embeddings.
  }
  \label{tab:fixed_token}
  \centering
  \setlength{\tabcolsep}{4pt}
  \begin{tabular}{c|cc|cc|cc|cc|c|c}
    \thickhline
    \multirow{2}{*}{Method} & \multicolumn{2}{c|}{ScanRefer} & \multicolumn{2}{c|}{Multi3DRefer} & \multicolumn{2}{c|}{Scan2Cap} & \multicolumn{2}{c|}{ScanQA} & SQA3D & \multirow{2}{*}{Avg.} \\
     & Acc@0.25 & Acc@0.5 & F1@0.25 & F1@0.5 & B-4@0.5 & C@0.5 & CIDEr & EM & EM & \\
    \hline \hline
    \rowcolor{title_gray}
    \multicolumn{11}{c}{\textit{8 Frames, All 1568 Tokens}} \\
    \hline
    Video-3D LLM$\dagger$~\cite{Video3DLLM} & 52.1 & 46.3 & 51.5 & 47.0 & 39.9 & 76.0 & 99.2 & 28.0 & 58.0 & 100\% \\
    \hline
    \rowcolor{title_gray}
    \multicolumn{11}{c}{\textit{16 Frames, Retain 1568 Tokens} \textcolor{blue}{($\downarrow$50\%)}} \\
    \hline
    Geo3DPruner~(Ours) & 58.2 & 51.9 & 56.9 & 51.8 & 41.6 & 85.3 & 100.8 & 29.2 & 59.5 & 108\% \\
    \hline
    \rowcolor{title_gray}
    \multicolumn{11}{c}{\textit{32 Frames, Retain 1568 Tokens} \textcolor{blue}{($\downarrow$75\%)}} \\
    \hline
    Geo3DPruner~(Ours) & 60.5 & 53.9 & 58.0 & 52.9 & 42.3 & 86.8 & 98.8 & 28.3 & 58.7 & 109\% \\
    \hline
  \end{tabular}
\end{table*}

\subsection{Comparison under Fixed Token Budget}
We further conduct a comparison under a fixed token budget of 1568 tokens. The baseline processes 8 frames without pruning, while our method utilizes more frames~(16 and 32) combined with pruning ratios of 50\% and 75\%, respectively, to maintain the same number of visual tokens.
As shown in Tab.~\ref{tab:fixed_token}, Geo3DPruner achieves substantially better performance on average, improving from 100\% to 108\% and 109\% when using 16 and 32 frames, respectively.
These results highlight a key advantage of our method, as it enables the model to process a larger number of frames while selectively retaining the most informative tokens. By leveraging geometry-guided pruning to remove cross-view redundancy, our method effectively increases the diversity of spatial observations under the same token budget, leading to more complete and robust 3D scene representations.

\begin{figure}[t]
  \centering
  \includegraphics[width=1.0\linewidth]{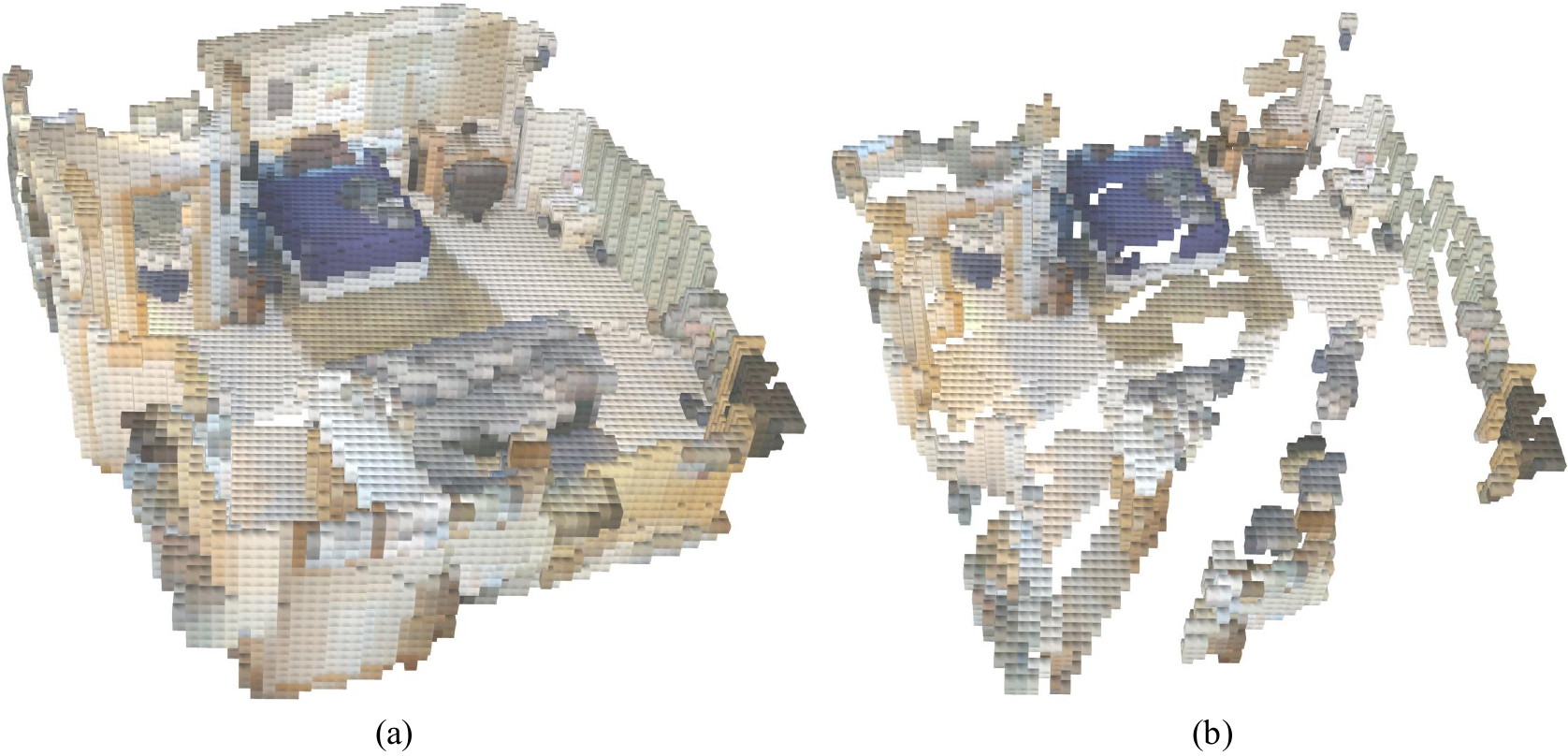}
  \caption{
  \textbf{Voxel-level visualization before and after pruning.} (a) Visualization of the original voxelized scene, where voxels densely cover both foreground objects and background regions. (b) Visualization after applying Geo3DPruner, where redundant voxels are largely removed while object-related and structurally important voxels are preserved. 
  }
  \label{fig:voxel_scene}
\end{figure}

\section{Voxel-level Visualization}
To provide a more intuitive understanding of the proposed voxel-level pruning strategy, we visualize the 3D voxel representation of the scene before and after pruning. 
As shown in Fig.~\ref{fig:voxel_scene}, the original voxel grid contains a large number of voxels distributed across both foreground objects and background regions. After applying Geo3DPruner, redundant voxels are effectively removed, while voxels corresponding to object-centric and structurally important regions are well preserved. This comparison clearly demonstrates that our method significantly reduces voxel redundancy while maintaining the overall spatial structure of the scene.